\documentclass[letterpaper, 10 pt, conference]{ieeeconf}  

\IEEEoverridecommandlockouts                              

\usepackage{graphicx} 
\usepackage{amsmath} 
\usepackage{amssymb}  
\usepackage{booktabs}
\usepackage[colorlinks, linkcolor=cyan, citecolor=green,]{hyperref}
\newcommand{\bestresult}[1]{\textcolor{red}{\textbf{#1}}}
\newcommand{\sbresult}[1]{\textcolor{blue}{#1}}

\title{\LARGE \bf
PMM-Net: Single-stage Multi-agent Trajectory Prediction with Patching-based Embedding and Explicit Modal Modulation
}

\author{Liu Huajian, Dong Wei, Fan Kunpeng, Wang Chao and Gao Yongzhuo$^{\ast}$
\thanks{All authors are with the State Key Laboratory of Robotics and System, Harbin Institute of Technology, Harbin, China. ($\ast$: Corresponding author)
        {\tt\small hjliu@stu.hit.edu.cn, gaoyongzhuo@hit.edu.cn}}%
}

\begin{document}

\maketitle
\thispagestyle{empty}
\pagestyle{empty}

\begin{abstract}

Analyzing and forecasting trajectories of agents like pedestrians plays a pivotal role for embodied intelligent applications. The inherent indeterminacy of human behavior and complex social interaction among a rich variety of agents make this task more challenging than common time-series forecasting. In this letter, we aim to explore a distinct formulation for multi-agent trajectory prediction framework. Specifically, we proposed a patching-based temporal feature extraction module and a graph-based social feature extraction module, enabling  effective feature extraction and cross-scenario generalization. Moreover, we reassess the role of social interaction and present a novel method based on explicit modality modulation to integrate temporal and social features, thereby constructing an efficient single-stage inference pipeline. Results on public benchmark datasets demonstrate the superior performance of our model compared with the state-of-the-art methods. The code is available at:  github.com/TIB-K330/pmm-net.

\end{abstract}

\section{INTRODUCTION}

Multi-agent trajectory prediction (MATP) problem aims to predict the future trajectories for one or multiple interacting agents conditioned on their past movements. It serves as a critical component for autonomous navigation tasks in dynamic environments, providing essential perception information for planners during decision-making and obstacle avoidance \cite{li2023marc}. Compared to conventional time-series forecasting (TSF) problem, MATP places greater emphasis on the accuracy of short-term predictions rather than long-term trends, while also requiring high real-time performance. In addition, MATP differs from TSF in two key challenges: handling complex interactions among agents, and addressing the inherent indeterminacy of human behavior, which leads to the multi-modal nature of future states.  

Existing mainstream approaches have commonly adopted Transformer-based architectures that are widely used in NLP domain \cite{vaswani2017attention}, leveraging their remarkable capability of extracting semantic correlations among elements in sequences. However, recent studies \cite{zeng2023transformers} have demonstrated that such architectures are redundant and inefficient for time-series forecasting. This is because self-attention is permutation-invariant, requiring positional encoding or other methods to preserve temporal information. The semantic information contained in individual trajectory points is too sparse, leading to a loss of temporal relationships between tokens in deeper network layers, which ultimately results in performance degradation of the entire model. Therefore, it is necessary to explore more effective feature extraction methods with lower computational complexity.

For multi-modal prediction, existing approaches often rely on generative models or employ goal-conditioned frameworks. Generative models represent multi-modality using a latent variable, with typical examples including generative adversarial networks (GANs), conditional variational auto-encoders (CVAEs), and more recent frameworks based on probabilistic diffusion models (DPMs). Goal-conditioned methods usually adopt a two-stage training framework, predicting candidate endpoints then guiding the the regression of the future trajectory. However, generative models are often inefficient and struggle to meet the high real-time demands of autonomous robotic tasks. On the other hand, the two-stage architecture used in goal-conditioned approaches can lead to unstable training or produce unnatural trajectories.

To address these critical gaps in MATP problem, we present a novel approach in this letter. Inspired by recent advances in TSF \cite{nie2023a}, we propose a patching-based temporal feature extraction module. By incorporating sub-series-level patches, this module enhances locality and captures comprehensive semantic information that is unavailable at the point level. In addition, we design a graph-based social feature extraction module that ensures translational and rotational invariance under series-level normalization, improving the model's generalization ability across different scenarios. Furthermore, we construct a novel single-stage framework through a cross-attention-based modality modulation using social information, which decouples temporal and social features, thereby reducing the overall computational complexity. Extensive experiments demonstrate that our method accurately predicts plausible future trajectories with multi-modality, achieving state-of-the-art results on Stanford Drone and ETH/UCY datasets. 

The main contributions of this paper can be summarized as follows: (1) We propose a sliding-window patching-based input embedding scheme that addresses the limitation of semantic correlation loss caused by directly embedding 2D points as tokens, enabling effective temporal feature extraction for agents; (2) We propose an novel social feature extraction method that utilizes an inverted attention mechanism among agents, achieving low computational complexity while ensuring translational and rotational invariance of the scene; (3) We design a distinct single-stage framework named PMM-Net that employs a cross-attention-based explicit modality modulation to achieve multi-modal prediction for MATP,  while more effectively supporting subsequent decision-making and planning tasks.
\section{RELATED WORKS}
\subsection{Social and Temporal Feature Representation}
For general learning-based TSF tasks including MATP, the essence lies in extracting temporal features through observing historical sequences, thereby constructing an implicit model. Specifically, MATP task requires modeling social interactions by aggregating the influences of neighboring agents.

\textbf{Temporal Features}: Similar to sequence modeling, common MATP approaches often directly adopt RNNs such as LSTM and GRU to capture the temporal features of observed trajectories. However, these methods struggle to effectively model long-term temporal correlations. Specially, \cite{lv2023skgacn} uses a temporal convolutional network to extract temporal features, which can process spatio-temporal interaction features in parallel and reduce computational complexity. \cite{wong2022view} models and forecasts trajectories from the spectral domain, employing the Discrete Fourier Transform to obtain trajectory spectrums to capture agents' detailed motion preferences across different frequency scales. \cite{xu2022remember, marchetti2024smemo} exploits an external addressable memory to explicitly store representative instances in the training set. By recalling related instances from memory instead of summarizing the entire history of observed inputs into a single internal latent state, these methods overcome the information loss issue of RNN. 

With the invention of Transformers and positional encoding, many works start to adopt self-attention mechanism  for sequence modeling due to their strong ability to capture long-range dependencies. However, recent studies indicate that directly using the self-attention mechanism does not necessarily improve, and may even degrade, the performance of TSF. While works in TSF, such as \cite{nie2023a, donghao2024moderntcn}, have explored various modifications to mitigate the performance degradation of Transformer-based methods in temporal prediction, no comprehensive solution for temporal feature extraction in MATP has yet been developed.

\textbf{Social Features}: Social interaction is a unique type of implicit information in MATP. Early works often adopted or improved upon the social pooling mechanism proposed by Alahi et al. to aggregate the hidden states of neighboring agents within a certain distance threshold. Grid-based methods, such as \cite{salzmann2020trajectron++}, were subsequently proposed to explore additional simple rules to enhance their capacity. Specially, \cite{wong2024socialcircle} introduced a grid-based approach using polar coordinates which combines the interpretability of model-based approaches with data-driven enhancements, reducing the reliance on complex network structures and high-quality training data. However, these pooling-based methods assume that relative distance between agents is the key factor in interaction modeling, which is not always accurate. Besides, SMEMO \cite{marchetti2024smemo}, building on \cite{xu2022remember}, exploits the external memory as a shared workspace to reason about social interactions between multiple agents.

Due to inherent structural similarities, spatio-temporal graphs are widely used to model social scenarios. A representative example is \cite{salzmann2020trajectron++}, which extracts edge features from neighboring agents through element-wise summation and LSTM, followed by an additive attention mechanism to aggregate the edge features connected to the same node, constructing a social influence vector. Agentformer \cite{yuan2021agentformer} designs a special agent-aware attention mechanism, allowing direct feature interaction across time and agents. Methods such as \cite{zhao2021you, zhou2022dynamic, lv2023skgacn, mao2023leapfrog, chen2023goal, chen2023vnagt} directly employ GAT \cite{velivckovic2018graph} or its variants to simulate interactions as the edges between different nodes. However, these methods rely on refining temporal features, and the stacking of spatio-temporal attention significantly impacts the model's computational efficiency, while limiting the use of generalized normalization methods that could effectively enhance model performance.
\subsection{Multi-modal Trajectory Prediction}
The multi-modal prediction task is non-trivial, as a single input can correspond to multiple outputs. To address this issue, existing methods can generally be categorized into two major types: generative methods, which represent the multiple possible future trajectories as a probability distribution, and another type that predicts several possible goals, using them to guide the regression of the trajectories.

For generative methods, early pioneering work Social-GAN \cite{gupta2018social} employed GANs to generate multi-modal trajectories. However, GANs are difficult to train and can suffer from mode collapse. Approaches such as \cite{salzmann2020trajectron++, yuan2021agentformer, zhou2022dynamic} explicitly handle multi-modality by leveraging the CVAE latent variable framework, but this approach can sometimes produce unnatural or unrealistic results. The diffusion model, a recent advanced generative framework inspired by non-equilibrium thermodynamics, was employed by Gu et al. in MID \cite{gu2022stochastic} to model the indeterminacy of human behavior. Following MID, LED \cite{mao2023leapfrog} incorporated a leapfrog initializer to skip many denoising steps to reduce inference time. 

For goal-conditioned approaches, TNT \cite{zhao2021tnt} was the first to adopt candidate goal prediction to assist a deterministic model in generating diverse outputs in parallel. PCCSNet \cite{sun2021three} utilizes a deep clustering process to obtain multiple modality representations. \cite{zhao2021you} employs similarity search to match observed trajectories with training data stored in an expert repository, thereby obtaining target candidates. SGNet \cite{wang2022stepwise} estimates and utilizes goals at multiple temporal scales to better capture the potentially varying intentions of pedestrians. SRGAT \cite{chen2023goal} further considers the mutual influences among multi-modal target points for different agents. However, as explained earlier, generative methods are inefficient, and goal-conditioned methods that employ two-stage architectures require additional hyper-parameter tuning, which can lead to an unstable training process.
\section{METHODOLOGY}
\subsection{Problem Formulation}
The multi-pedestrian trajectory prediction problem can be formulated as following: given $N$ historical trajectories $\mathcal{X}$ in a 2D scenario $\mathcal{W} \subseteq \mathbb{R}^2$ of length $T^{\prime} \in \mathbb{N}^*$, the goal is to generate plausible $K$-modal future trajectories of a specific length $T \in \mathbb{N}^*$ for each pedestrian based on all prior information. Each input historical trajectory can be denoted as $\mathbf{X}^i = [\mathbf{p}_{-T^{\prime}+1}^i, \mathbf{p}_{-T^{\prime}+2}^i, \dots, \mathbf{p}_0^i], \forall i \in \{1, 2, \dots, N\}$, and the corresponding predicted future trajectories can be written as $\hat{\mathbf{Y}}^i = [\hat{\mathbf{p}}_1^i, \hat{\mathbf{p}}_2^i, \dots, \hat{\mathbf{p}}_T^i]$. In this research, we focus solely on uni-modal input, meaning we consider only the trajectory information within the same scene and ignore map or other environmental context.

\subsection{Patching-based Temporal Feature Extraction}
\label{subsec:temporal}
\begin{figure}[htbp]
    \centering
    \includegraphics[width=1.0\columnwidth]{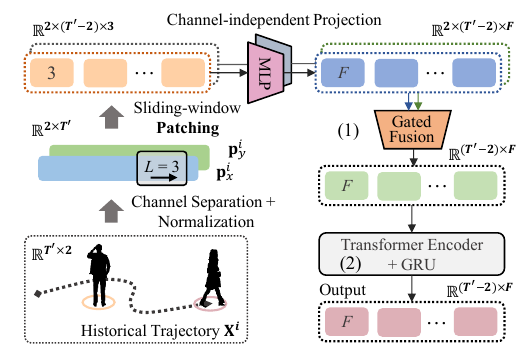}
    \caption{Illustration of the patching-based temporal feature extraction module, where the input consists of the observed historical trajectories, and the output is the encoded tokens.\label{fig:temporal}}
\end{figure}

The original uni-modal input for the trajectory prediction problem of a specific agent $i$ is a two-channel sequence of 2D position coordinates $\mathbf{X}^i \in \mathbb{R}^{T^{\prime} \times 2}$. Certain existing methods \cite{chen2023goal} use techniques like differencing to obtain pseudo-velocity $\bar{\mathbf{v}}_i$ or pseudo-acceleration $\bar{\mathbf{a}}_i$ as input for data augmentation, but this method has significant limitations in terms of scenario generalization. Our patching-based module further delves into this concept by utilizing a learnable approach to capture high-dimensional kinematic information. As illustrated Fig. \ref{fig:temporal}, for the two series channels $\mathbf{p}_x^i$ and $\mathbf{p}_y^i \in \mathbb{R}^{T^{\prime}}$, we applay two channel-independent sliding-window MLPs of length $P = 3$ with ReLU activation to map them into high-dimensional latent state spaces, denoted as $\mathbf{Z}_x$ and $\mathbf{Z}_y \in \mathbb{R}^{(T^{\prime} - 2) \times F}$.We then use a gated fusion, as shown in (\ref{eq:gated_fusion}), to merge the two channels, resulting in $T^{\prime} - 2$ input tokens, denoted as $\mathbf{Z} \in \mathbb{R}^{(T^{\prime} - 2) \times F}$.

\begin{equation}
    \left\{ \begin{aligned} &\boldsymbol{\omega} = \text{Sigmoid}(\mathbf{W}[\mathbf{Z}_x \parallel \mathbf{Z}_y] + \mathbf{b}) \\ &\mathbf{Z} = \boldsymbol{\omega} \odot \mathbf{Z}_x + (\mathbf{1} - \boldsymbol{\omega}) \odot \mathbf{Z}_y \end{aligned} \right.,
    \label{eq:gated_fusion}
\end{equation}
where the operator pair $(\mathbf{W}, \mathbf{b})$ denotes a trainable linear projection with bias\footnote{To avoid overly complex notation, we will use unmarked symbols to represent the learnable linear transformations throughout the remainder of this paper. Unless otherwise specified, these parameters do NOT share weights.} and operator $\parallel$ denotes concatenation. Based on the aforementioned embedding method, we design an encoder that combines self-attention with GRU to extract temporal feature between tokens $\mathbf{Z}$. Compared to a pure Transformer-based architecture, the proposed method has advantages in few-shot settings. We use a vanilla 3-layer multi-head Transformer encoder with additive positional encoding, which we won't elaborate on here. Its output, $\mathbf{Z}^{\prime}$, retains the same dimensions as $\mathbf{Z}$. Finally, the GRU \cite{cho2014properties} is employed to enhance the temporal relationships between the tokens $\mathbf{z}^t \in \mathbf{Z}^{\prime}$ as:
\begin{equation}
    \tilde{\mathbf{z}}^t, \mathbf{c}^t = \text{GRU}(\mathbf{z}^t, \mathbf{c}^{t-1}),
\end{equation}
where $\mathbf{c}^t$ denotes the hidden state output by the GRU at timestamp $t$. We concatenate all the outputs $\tilde{\mathbf{z}}^t$ to form the agent's final temporal features, denoted as $\tilde{\mathbf{Z}}$.

\subsection{Graph-based Social Interaction Representation}
\label{subsec:graph}
To effectively aggregate the social interaction information of neighboring agents in a scene while ensuring rotational and translational invariance, we propose a novel directed graph-based scene normalization method along with a graph neural network (GNN) based feature extraction module. For the historical trajectories of all agents in the same scene $\mathcal{X}$, we translate them to have $t = 0$ as the origin, while defining the features of directed edges based on the relative positional relationships between agents. Specifically, for a central agent $i$, the edge features in the graph, pointing toward this agent, represent the relative positional relationships between neighboring agents $j \in \mathcal{N}_i$ and $i$. These relationships are defined using a polar coordinate system, where the direction of the pseudo-velocity vector $\bar{\mathbf{v}}_i$ of the central agent serves as the zero-angle reference, as follows:

\begin{equation}
    \left\{ \begin{aligned} &\mathbf{d}_{ij} = (p_x^i - p_x^j, p_y^i - p_y^j) \\ &\cos\theta_{ij} = \frac{\langle\mathbf{d}_{ij}, \bar{\mathbf{v}}_i\rangle}{\|\mathbf{d}_{ij}\| \| \bar{\mathbf{v}}_i \|} \end{aligned} \right.
\end{equation}

To ensure numerical stability, we use the cosine of the angle $\theta_{ij}$ instead of the angle itself. Further, we employ an MLP with ReLU activation to extract high-dimensional features for the edges, which is defined as follows:

\begin{equation}
    \mathbf{e}_{ij} = \text{MLP}(\mathbf{d}_{ij} \parallel \cos \theta_{ij})
\end{equation}

\begin{figure}[htbp]
    \centering
    \includegraphics[width=1.0\columnwidth]{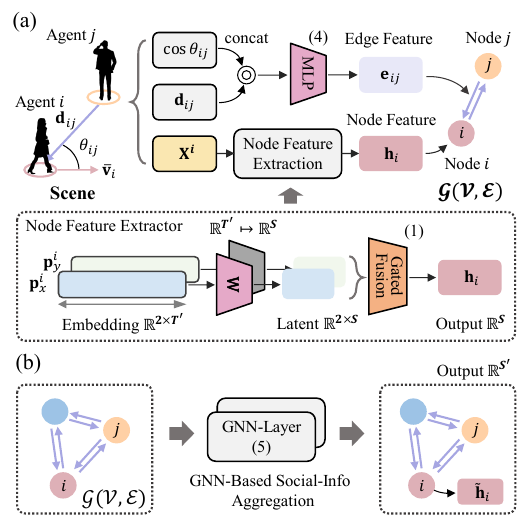}
    \caption{Illustration of the graph-based social feature extraction module: a) Graph representation of a specific crowd scenario, where $\mathcal{V}$ denotes the set of all nodes and $\mathcal{E}$ denotes the set of all edges; b) GNN-based social information aggregation.\label{fig:social}}
\end{figure}

Inspired by \cite{liu2024itransformer}, we use an inverted channel-independent encoding method to extract node state features, as shown in Fig. \ref{fig:social}. Specifically, the two channels of the node's historical trajectory, $\mathbf{p}_x$ and $\mathbf{p}_y$, are first linearly projected to higher dimensions, then fused via a gated mechanism to obtain the node feature $\mathbf{h} \in \mathbb{R}^S$. We then apply a GNN to extract features from the directed graph. Based on \cite{velivckovic2018graph}, we define the message passing function of the GNN as follows:

\begin{equation}
    \left\{ \begin{aligned} &\mathbf{u}_{ij} = \text{LeakyReLU}\left(\mathbf{W}[\mathbf{h}_i \parallel \mathbf{h}_j \parallel \mathbf{e}_{ij}] + \mathbf{b}\right) \\
    &\alpha_{ij} = \text{Softmax}_j(\mathbf{u}_{ij}) = \frac{\exp(\mathbf{u}_{ij})}{\sum_{k \in \mathcal{N}_i} \exp(\mathbf{u}_{ik})} \\
    &\tilde{\mathbf{h}}_i = \text{PReLU}\left(\sum\limits_{j \in \mathcal{N}_i} \alpha_{ij} [\mathbf{W}\mathbf{h}_j + \mathbf{b}]\right)
    \end{aligned} \right.,
\end{equation}
where the operator pair $(\mathbf{W}, \mathbf{b})$ represents a learnable linear transformation. To stabilize the learning process of self-attention, we found it beneficial to extend our mechanism by employing multi-head attention, similar to \cite{vaswani2017attention}. The resulting $\tilde{\mathbf{h}}_i \in \mathbb{R}^{S'}$ is used as the agent's social feature.

\subsection{Multi-modal Linear Projection with Social Refinement}
Inspired by \cite{carion2020end}, we propose a novel single-stage framework to handle the multi-modal prediction requirement in MATP, while introducing a new perspective on the role of social information. By employing multi-modal linear projections, we map the temporal feature of agents into a fixed number $K$ of modalities, and instead of predicting end-points or Gaussian probability distributions of future trajectories, as in existing approaches, we assign a discrete probability score to each trajectory through an independent explicit scoring head. Formally, we have:
\begin{equation}
    \mathbf{f}_i^k = \text{MLP}_k(\tilde{\mathbf{Z}}_i), k \in \{1, \ldots , K\} := \mathcal{K},
\end{equation}
where $\text{MLP}_k$ represents the k-th modality-specific MLP with dedicated parameters. $\mathbf{f}_i^k \in \mathbb{R}^H$ serves as the high-dimensional latent variable for each potential modality.

\begin{figure}[htbp]
    \centering
    \includegraphics[width=1.0\columnwidth]{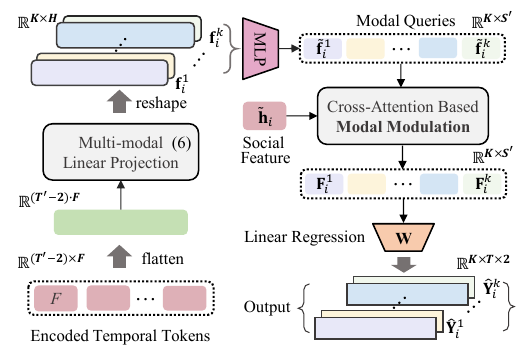}
    \caption{Illustration of the single-stage multi-modal prediction framework.\label{fig:projection}}
\end{figure}

As illustrated in Fig. \ref{fig:projection}, our framework can be loosely regarded as a reversed encoder-decoder architecture. The social feature $\tilde{\mathbf{h}}_i$, described in Section \ref{subsec:graph}, is treated as the output of the encoder, while $\{\mathbf{f}_i^k\ | \forall k \in \mathcal{K}\}$ serves as the input to the decoder. We use a shared-weight MLP with a PReLU activation function to map $\mathbf{f}_i^k$ into the same dimension as $\tilde{\mathbf{h}}_i$, denoted as $\tilde{\mathbf{f}}_i^k$. Different from the standard Transformer used in Section \ref{subsec:temporal}, in the modality modulation module, we define the cross-attention between $\tilde{\mathbf{h}}_i$ and $\tilde{\mathbf{f}}_i^k$ as follows:
\begin{equation}
    \mathbf{A}_i^k := \text{Attention}(\tilde{\mathbf{h}}_i, \tilde{\mathbf{f}}_i^k) = \text{Softmax}\left( \cfrac{\langle \mathbf{Q}_i, \mathbf{K}_i^k\rangle}{\sqrt{d}} \right) \mathbf{V}_i^k
\end{equation}
where $\mathbf{Q}_i, \mathbf{K}_i^k, \mathbf{V}_i^k = \mathbf{W}^Q\tilde{\mathbf{h}}_i, \mathbf{W}^K \tilde{\mathbf{f}}_i^k,  \mathbf{W}^V \tilde{\mathbf{f}}_i^k$, with each $\mathbf{W}$ being a learnable weight. Based on this, similar to the original block of Transformer, we define the output of the decoder as:
\begin{equation}
    \left\{ \begin{aligned} &\tilde{\mathbf{A}}_i^k = \text{LayerNorm}(\tilde{\mathbf{f}}_i^k + \mathbf{A}_i^k) \\
    &\mathbf{F}_i^k = \text{LayerNorm}(\tilde{\mathbf{A}}_i^k + \text{FFN}(\tilde{\mathbf{A}}_i^k))
    \end{aligned} \right.,
\end{equation}
where the operator $\text{FFN}(\cdot)$ represents a feed-forward network. For each $\mathbf{F}_i^k \in \mathbb{R}^{S'}$, we employ a linear mapping with learnable weights shared across modalities, defined as $\hat{\mathbf{Y}}_i^k = \mathbf{W}\mathbf{F}_i^k + \mathbf{b}$, to regress the final future trajectories. Similarly, we define the scoring head using an additional linear layer followed by a softmax function:
\begin{equation}
  \mathbf{P}_i = \text{Softmax}_k (\mathbf{W}\mathbf{F}_i^k + \mathbf{b}),
\end{equation}
where $\mathbf{P}_i$ represents a normalized score, indicating the explicit probability of each modality.

\subsection{Training Objective}
The proposed framework incorporates two parallel loss components concerning trajectory prediction and classification to capture multi-modal futures. Given the ground truth trajectory $\{ \mathbf{p}_t^i\}^T_{t=1}$ of a specific agent $i$, the trajectory loss $\mathcal{L}_{\text{traj}}$ can be defined by calculating  the mean of the per-step difference only for the most accurate trajectory over $K$ predicted modalities $\hat{\mathbf{Y}}_i^k$:

\begin{equation}
    \mathcal{L}_{\text{traj}} = \frac{1}{T}\min {}_{k=1}^K \sum_{t=1}^T\| \hat{\mathbf{p}}^{i,k}_t - \mathbf{p}^i_t\|^2
\end{equation}

Classification loss $\mathcal{L}_{\text{cls}}$ is the Binary Cross Entropy Loss applied to the assigned probabilities $\mathbf{P}_i$ of $K$ trajectories, where the ground truth probability of the closest trajectory is set to 1 and the others to 0:
\begin{equation}
    \mathcal{L}_{\text{cls}} = \mathcal{L}_{\text{BCE}}(\mathbf{P}_i, \mathbf{y}_i),
\end{equation}
where $\mathbf{y}_i$ denotes the ground truth probabilities assigned. Overall, our training objective is to minimize the sum of these two losses, denoted as $\mathcal{L} = \mathcal{L}_{\text{traj}} + \mathcal{L}_{\text{cls}}$.

\section{EXPERIMENTS}
\subsection{Experimental Setup}
\textbf{Datasets}: We conducted a quantitative evaluation of the proposed method based on two well-established public datasets including ETH/UCY and Stanford Drone Dataset (SDD). The ETH/UCY dataset group is the most commonly used public dataset for pedestrian trajectory prediction tasks and consists of five diverse sub-scenes. These scenes are captured in unconstrained environments with few obstacles and rich interactions. Specifically, the \textit{ETH} and \textit{Hotel} scenes are recorded in the city center of Zurich, while the \textit{Univ}, \textit{Zara1}, and \textit{Zara2} scenes are captured on the campus of the University of California, Berkeley. This dataset provides ground-truth trajectories in a world coordinate system, with a sampling frequency of 2.5 Hz.

The SDD dataset is a large-scale MATP dataset captured from a bird's-eye view, containing 20 scenes and 6 different categories of agents, covering various outdoor locations, including intersections and roundabouts, thus providing a realistic representation of real-world scenarios. SDD provides ground-truth bounding-box labeled in pixel units, with a frame rate of 30 Hz. We employ a downsampling preprocessing method, setting the time interval $\Delta t$ of the position sequences to 0.4 seconds.

\textbf{Metrics}: We adopted the widely-used evaluation metrics, reporting the minimum average displacement error ADE$_K$ and final displacement error FDE$_K$ of $K = 20$ trajectories generated for each agent compare to the ground truth, defined as: $\text{ADE}_K := \frac{1}{T}\min_{k=1}^K \sum_{t=1}^T\| \hat{\mathbf{p}}_t^{i, k} - \mathbf{p}_t^i \|^2$, $\text{FDE}_K := \min_{k=1}^K\|  \hat{\mathbf{p}}_T^{i, k} - \mathbf{p}_T^i \|^2$. Here, $\hat{\mathbf{p}}_t^{i, k}$denotes the predicted future position of agent $i$ at time $t$ in the $k$-th modal and $\mathbf{p}_t^i$ is the corresponding ground truth. Similarly, we followed established standards, using $T^\prime = 8$ frames (3.2 seconds) of historical trajectories to predict $T = 12$ frames (4.8 seconds) of future trajectories. For the ETH/UCY dataset, since there is no standard test set, we followed the leave-one-out strategy for evaluation, as in prior work. Specifically, we trained our model on four scenes and tested on the remaining one.

\textbf{Implementation Details}: We set the hidden dimension $F$ for temporal features to 64, the dimensions $S$ and $S^\prime$ for social features to 256, and the dimension $H$ of the latent modality variable to 1024 based on experimental results and empirical insights. In addition, we set the number of heads to 4 for all modules utilizing multi-head attention. We train our models using Adam optimizer, with an initial learning rate of 5e-4 and a cosine annealing learning rate schedule. Following \cite{chen2023goal}, random rotation was applied for data augmentation during training. For the ETH/UCY dataset, we set the batch size to 32, the maximum number of epochs to 300, and the maximum interaction distance between two agents to 10 meters. For the SDD dataset, we set the batch size to 128, the maximum number of epoch to 200 and the corresponding maximum distance is set to 200 pixels. The proposed model is implemented based on the open-source deep learning framework PyTorch (DGL for graph neural networks) and and all experiments were conducted on a single NVIDIA RTX4070Ti GPU without any pre-training.
\subsection{Quantitative Results}
\textbf{ETH/UCY}: For the ETH/UCY dataset group, we select six representative SOTA methods from the past two years as baselines for comparison. Among these, \textbf{SGNet-ED} \cite{wang2022stepwise} and \textbf{SRGAT} \cite{chen2023goal} represent goal-conditioned methods; \textbf{MID} \cite{gu2022stochastic} and \textbf{LED} \cite{mao2023leapfrog} represent diffusion-based approaches, with the latter optimizing inference speed over the former; \textbf{DACG} \cite{zhou2022dynamic} combines a CVAE with GAN and incorporates an improved GAT for social feature extraction, similar to the proposed method; Specically, \textbf{SMEMO} \cite{marchetti2024smemo} serves as a representative of the memory-augmented neural network architecture. In addition, we also include six widely influential traditional methods, which do not require further elaboration here.

We summarize the test results on the ETH/UCY dataset in Table \ref{tab:eth-ucy}, highlighting the best performance in bold-red and the second best in blue. Results reproduced based on the open-source code are denoted by $\dagger$. It can be intuitively observed that our proposed model outperforms all baseline methods in most cases, achieving the highest average ADE$_{20}$/FDE$_{20}$ and the best or second best to the best performance across the all five scenarios. Specifically, compared with current SOTA method SRGAT, which adopts a goal-conditioned multi-modal prediction approach, the proposed method reduces the average ADE$_{20}$/FDE$_{20}$ from 0.19/0.31 to 0.17/0.28, achieving 10.5\%/9.6\% improvement. Compared with current SOTA method LED, which utilizes a probabilistic model for multi-modality, the proposed method reduces the average ADE$_{20}$/FDE$_{20}$ from 0.21/0.33 to 0.17/0.28, achieving 19.0\%/15.2\% improvement. 

\begin{table*}[htbp]
    \caption{Quantitative comparison results on the ETH/UCY dataset with Best-of-20 strategy in ADE/FDE metric (The lower the better).\label{tab:eth-ucy}}
    \centering
    \begin{tabular}{l || l l l l l l}
        \toprule
        Model & ETH & Hotel & Univ & Zara1 & Zara2 & Average \\
        \midrule
        Social-GAN \cite{gupta2018social} \; \textcolor{blue}{(CVPR'18)} & 0.81/1.52 & 0.72/1.61 & 0.60/1.26 & 0.34/0.69 & 0.42/0.84 & 0.58/1.18 \\
        PECNet \cite{mangalam2020not} \; \textcolor{blue}{(ECCV'20)} & 0.54/0.87 & 0.18/0.24 & 0.35/0.60 & 0.22/0.39 & 0.17/0.30 & 0.29/0.48 \\
        Trajectron++ \cite{salzmann2020trajectron++} \; \textcolor{blue}{(ECCV'20)} & 0.39/0.83 & 0.12/0.21 & \bestresult{0.20}/0.44 & \sbresult{0.15}/0.33 & \sbresult{0.11}/0.25 & 0.19/0.41 \\
        PCCSNET \cite{sun2021three} \; \textcolor{blue}{(ICCV'21)} & 0.28/0.54 & \sbresult{0.11}/0.19 & 0.29/0.60 & 0.21/0.44 & 0.15/0.34 & 0.21/0.42 \\
        Agentformer \cite{yuan2021agentformer} \; \textcolor{blue}{(ICCV'21)} & 0.45/0.75 & 0.14/0.22 & 0.25/0.45 & 0.18/0.30 & 0.14/0.24 & 0.23/0.39 \\
        $\dagger$ Expert-Goals \cite{zhao2021you} \; \textcolor{blue}{(ICCV'21)} & 0.37/0.65 & \sbresult{0.11}/\bestresult{0.15} & \bestresult{0.20}/0.44 & \sbresult{0.15}/0.31 & 0.12/0.25 & 0.19/0.36 \\
        SGNet-ED \cite{wang2022stepwise} \; \textcolor{blue}{(RA-L'22)} & 0.35/0.65 & 0.12/0.24 & \bestresult{0.20}/0.42 & \bestresult{0.12/0.24} & \bestresult{0.10/0.21} & \sbresult{0.18}/0.35 \\
        MID \cite{gu2022stochastic} \; \textcolor{blue}{(CVPR'22)} & 0.39/0.66 & 0.13/0.22 & 0.22/0.45 & 0.17/0.30 & 0.13/0.27 & 0.21/0.38 \\
        DACG \cite{zhou2022dynamic} \; \textcolor{blue}{(RA-L'23)} & 0.36/0.65 & 0.13/0.25 & \bestresult{0.20}/0.41 & 0.16/0.34 & 0.12/0.26 & 0.19/0.38 \\
        LED \cite{mao2023leapfrog} \; \textcolor{blue}{(CVPR'23)} & 0.39/0.58 & \sbresult{0.11}/0.17 & 0.26/0.43 & 0.18/\sbresult{0.26} & 0.13/\sbresult{0.22} & 0.21/0.33 \\
        SMEMO \cite{marchetti2024smemo} \; \textcolor{blue}{(T-PAMI'24)} & 0.39/0.59 & 0.14/0.20 & 0.23/0.41 & 0.19/0.32 & 0.15/0.26 & 0.22/0.35 \\
        $\dagger$ SRGAT \cite{chen2023goal} \; \textcolor{blue}{(RA-L'24)} & \sbresult{0.27/0.43} & \sbresult{0.11/0.16} & 0.22/\sbresult{0.39} & 0.21/0.35 & 0.12/\sbresult{0.22} & 0.19/\sbresult{0.31} \\
        \midrule
        \textbf{PMM-Net} \textcolor{blue}{(Proposed)} & \bestresult{0.26/0.38} & \bestresult{0.10/0.15} & \sbresult{0.21}/\bestresult{0.38} & 0.16/0.29 & 0.12/\bestresult{0.21} & \bestresult{0.17/0.28} \\
        \bottomrule
    \end{tabular}
\end{table*}

\textbf{SDD}: For the SDD dataset, we supplement the evaluation with a few additional baselines, including: Y-net \cite{mangalam2021goals}, Goal-SAR \cite{chiara2022goal}, V$^2$-Net-SC \cite{wong2024socialcircle} and SIM \cite{li2023synchronous}, as some approaches used in ETH/UCY dataset do not provide test results on the SDD dataset. Notably, unlike the ETH/UCY dataset, which primarily consists of unconstrained environments, the SDD dataset contains more terrain constraints. The first three additional baselines we introduced incorporate image-based map information, allowing for a more comprehensive comparison in our experiments.

We summarize the test results on the SDD dataset in Table \ref{tab:sdd} with the best performance highlighted in bold. ``T'' denotes the method only using the trajectory position information, and ``T + I'' denotes the method using both position and visual image information. Specifically, compared with current SOTA method  that uses only historical trajectories as input, SIM, the proposed method reduces the average ADE$_{20}$/FDE$_{20}$ from 7.40/11.39 to 6.30/10.34, achieving 14.9\%/9.2\% improvement. Compared with current SOTA method that uses additional RGB maps as context information, V$^2$-Net-SC, the proposed method reduces the average ADE$_{20}$/FDE$_{20}$ from 6.71/10.66 to 6.30/10.34, achieving 6.1\%/3.0\% improvement.

\begin{table}[htbp]
    \caption{Quantitative results on the Standford Drone Dataset with Best-of-20 strategy in ADE and FDE metric.\label{tab:sdd}}
    \begin{center}
        \begin{tabular}{l || l l l l}
            \toprule
            Model & Input & Samples $K$ & ADE & FDE \\
            \midrule
            Y-net \cite{mangalam2021goals} \textcolor{blue}{(2021)} & T + I & 20 & 8.97 & 14.61 \\
            Y-net + TITS & T + I & 10000 & 7.85 & 11.85 \\
            Goal-SAR \cite{chiara2022goal} \textcolor{blue}{(2022)} & T + I & 20 & 7.75 & 11.83 \\ 
            V$^2$-Net-SC \cite{wong2024socialcircle} \textcolor{blue}{(2024)} & T + I & 20 & 6.71 & 10.66 \\
            \midrule
            Social-GAN \cite{gupta2018social} \textcolor{blue}{(2018)} & T & 20 & 27.23 & 41.44 \\
            PECNet \cite{mangalam2020not} \textcolor{blue}{(2020)} & T & 20 & 9.96 & 15.88 \\
            PCCSNET \cite{sun2021three} \textcolor{blue}{(2021)} & T & 20 & 8.62 & 16.16 \\
            $\dagger$ Expert-Goals \cite{zhao2021you} \textcolor{blue}{(2021)} & T & 20 & 10.67 & 14.38 \\
            $\dagger$ Expert-Goals + GMM & T & 20$\times$20 & 7.65 & 14.38 \\
            MID \cite{gu2022stochastic} \textcolor{blue}{(2022)} & T & 20 & 7.91 & 14.50 \\ 
            LED \cite{mao2023leapfrog} \textcolor{blue}{(2023)} & T & 20 & 8.48 & 11.66 \\
            SMEMO \cite{marchetti2024smemo} \textcolor{blue}{(2024)} & T & 20 & 8.11 & 13.06 \\
            SRGAT \cite{chen2023goal} \textcolor{blue}{(2024)} & T & 20 & 7.86 & 13.13 \\
            SIM \cite{li2023synchronous} \textcolor{blue}{(2024)} & T & 20 & 7.40 & 11.39 \\
            \midrule
            \textbf{PMM-Net} \textcolor{blue}{(Proposed)} & T & 20 & \textbf{6.30} & \textbf{10.34} \\
            \bottomrule
        \end{tabular}
    \end{center}
\end{table}

\textbf{Model Complexity}: Besides prediction performance, we also assess model complexity using two metrics, the number of parameters (\#Param) and floating-point operations per second (FLOPS). The experimental results are summarized in Table \ref{tab:complexity}, we employ a limited number of baselines in this comparison as such data is rarely available directly from publicly published work. Due to our extensive use of linear projections to replace complex decoder structures, along with the decoupling of spatio-temporal feature encoding, it can be clearly seen that the proposed model achieves a significant performance improvement while maintaining a model complexity that is lower than, or at least comparable to, existing SOTA methods.
\begin{table}[htbp]
    \centering
    \setlength{\tabcolsep}{6mm}{
    \caption{Model complexity comparison.\label{tab:complexity}}
    \begin{tabular}{l | l l}
        \toprule
        Method & \#Param (M) & FLOPS (M) \\
        \midrule
        Social-GAN \cite{gupta2018social} & 3.907 & 30.053 \\ 
        PECNET \cite{mangalam2020not} & 2.241 & 25.882 \\
        VNGAT \cite{chen2023vnagt} & 0.225 & 6.097 \\
        SGCN \cite{shi2021sgcn} & 0.026 & 0.608 \\
        SRGAT \cite{chen2023goal} & 0.097 & 4.513 \\
        \midrule
        \textbf{PMM-Net} & 0.043 & 1.828 \\
        \bottomrule
    \end{tabular}}
\end{table}

\subsection{Ablation Studies}
In order to investigate the contribution of each submodule to the performance of the proposed method, we designed two targeted ablation model experiments. First, we replaced the patching-based temporal feature extraction module with a multi-layer perceptron (MLP), denoted as "w/o Patch". Secondly, we removed the social refinement module, resulting in a model that degrades to an encoder-only form, denoted as "w/o Social". We tested the performance of the two ablation models on the ETH/UCY and SDD datasets and summarized the experimental results in Table \ref{tab:ablation}.

\begin{table*}[htbp]
    \caption{Ablation Study Results on the ETH/UCY and SDD Datasets.\label{tab:ablation}}
    \centering
    \begin{tabular}{l || l l l l l l | l}
        \toprule
        Model & ETH & Hotel & Univ & Zara1 & Zara2 & ETH-UCY & SDD \\
        \midrule
        \textbf{PMM-Net} & 0.265/0.381 & \textbf{0.109/0.154} & \textbf{0.213/0.387} & \textbf{0.166/0.294} & \textbf{0.120/0.213} & \textbf{0.175/0.286} & \textbf{6.30/10.34} \\
        \midrule
        w/o Social & 0.265/0.395 & 0.117/0.169 & 0.216/0.394 & 0.173/0.310 & 0.123/0.217 & 0.179/0.297 & 6.36/10.54 \\
        w/o Patch & 0.275/0.418 & 0.114/0.163 & 0.215/0.393 & 0.175/0.315 & 0.125/0.222 & 0.181/0.302 & 6.42/10.70 \\ 
        Linear Proj. & \textbf{0.233/0.359} & 0.121/0.181 & 0.239/0.431 & 0.190/0.348 & 0.140/0.247 & 0.185/0.313 & N/A \\
        \bottomrule
    \end{tabular}
\end{table*}

It is evident that removing any component leads to inferior performance compared to the full model and the introduction of patching contributes significantly more to the improvement of model performance, while the effect of social interaction, although relatively smaller, is still noteworthy, aligning with the conclusions presented in \cite{makansi2022you}. We attribute the contribution of patching to performance enhancement to its ability to effectively capture comprehensive semantic information that is not available at the point level. Specifically, we employed a window length of 3 to leverage implicit local differentials to obtain higher-dimensional state information such as velocity and acceleration.
\subsection{Qualitative Evaluation}
\begin{figure*}[htbp]
    \centering
    \includegraphics[width=1.0\textwidth]{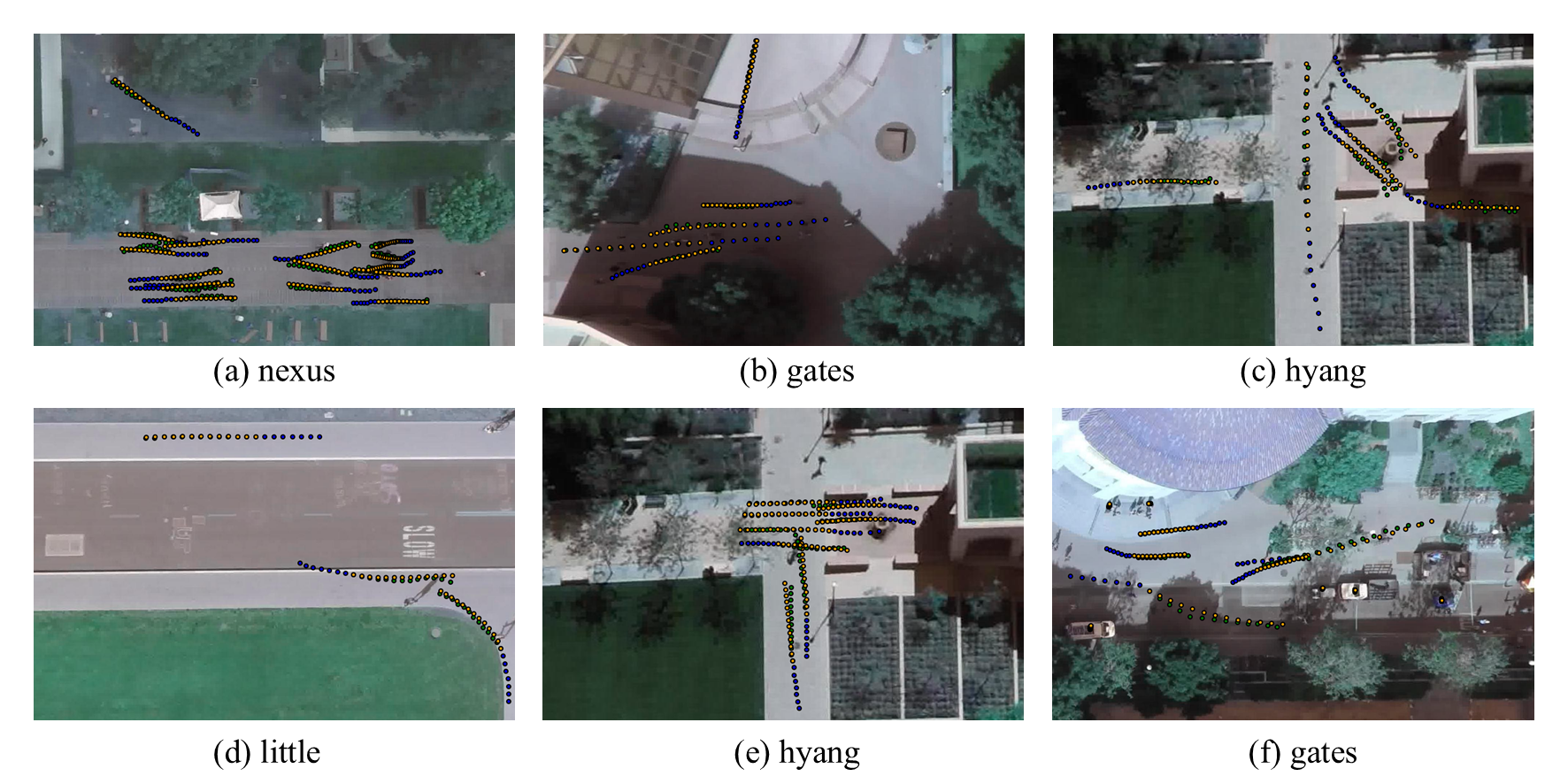}
    \caption{Qualitative results on SDD dataset with predicted trajectories are plotted in yellow, ground-truth plotted in green and observed historical trajectories depicted in blue. Each sub-figure is labeled with the name of its corresponding scene. The sampling interval between any two consecutive points on the same trajectory is 0.4 seconds, and its sparsity reflects variations in movement speed.\label{fig:visual}}
\end{figure*}

Besides the above quantitative experiments, we have selected several typical scenarios to visually demonstrate the proposed methods in a more intuitive manner, as shown in Fig. \ref{fig:visual}. From the figure, it can be intuitively observed that the proposed method is capable of handling various types of complex interaction scenarios, including but not limited to standing (f), turning (d)(f), grouping (a)(b)(e) and collision avoidance (c). In addition, our method demonstrates strong generalization adaptability for heterogeneous agents with varying movement speeds beyond just pedestrians.

\section{CONCLUSION AND DISCUSSION}
In this letter, we present PMM-Net, a novel single-stage framework to address MATP problem for robotic applications. Within this framework, we design an efficient patching-based temporal feature extraction module and a graph-based social feature extraction module. We reassess the role of social interaction in the MATP problem and innovatively devise a method based on explicit modality modulation to integrate temporal and social features, achieving effective feature extraction and cross-scenario generalization while ensuring low computational complexity. Experimental results demonstrate the superiority of our method which achieves state-of-the-art performance on both the SDD and ETH/UCY benchmarks while satisfying real-time inference needs for embodied intelligent agents.

We attribute the effectiveness of the temporal feature extraction module to the ability of local patches to capture high-order kinematic information of trajectory points through implicit differentiation, thereby enhancing the semantics in each embedded token. This enhancement enables subsequent self-attention layers to effectively capture temporal dependencies. In the social feature extraction module, we utilize a channel-independent series-global representation for the historical trajectories of neighboring agents, which can also be viewed as a special patch with a length equal to that of the historical trajectory, embedding the normalized historical trajectories into variate tokens. Although this approach results in a slight decrease in performance, it effectively reduces computational complexity in scenarios with a high number of interacting pedestrians, ensuring real-time performance.

Our experimental results demonstrate that effective feature extraction is a sufficient condition for achieving good prediction performance with limited training data. In contrast, overly complex networks, while appearing fancy, do not provide significant performance improvements and can actually reduce the model's efficiency. Our ablation experiments indicate that, in addition to the two proposed submodules, linear projection plays a crucial role in the overall performance of the framework, surpassing all decoder-based structures, including GRU, LSTM, and Transformer-based models. This finding aligns with recent conclusions from related work \cite{li2023revisiting}. Furthermore, the performance of generative decoder-based methods generally lags behind that of contemporaneous goal-conditioned methods and exhibits significantly higher computational complexity. We believe that such architectures are not well-suited for the MATP task with a fixed prediction length.

Our future will primarily focus on extending the proposed method to scenarios with heterogeneous features like urban environments \cite{aydemir2023adapt}, as well as integrating downstream decision-making and planning tasks to further explore the practical applications of explicit trajectory scoring and multi-modal prediction.





\section*{ACKNOWLEDGMENT}
The author(s) used ChatGPT for proofreading and manually reviewed the final manuscript.

\bibliographystyle{IEEEtran}
\bibliography{IEEEabrv, ref}

\end{document}